\documentclass{article}

\usepackage{microtype}
\usepackage{graphicx}
\usepackage{subfigure}
\usepackage{cite}
\usepackage{booktabs} % for professional tables

\usepackage{hyperref}

\usepackage{algorithm,algorithmic}

\usepackage[accepted]{icml2021}

\icmltitlerunning{Loss Landscape and the Lottery Ticket Hypothesis}

\begin{document}

\twocolumn[
\icmltitle{Visualizing the Loss Landscape of Winning Lottery Tickets}

\begin{icmlauthorlist}
\icmlauthor{Robert Bain}{osu}
\end{icmlauthorlist}

\icmlaffiliation{osu}{Oregon State University, Corvallis, Oregon, USA}

\icmlcorrespondingauthor{Robert Bain}{bainro@oregonstate.edu}

\icmlkeywords{Machine Learning, ICML}

\vskip 0.3in
]

\begin{abstract}
The underlying loss landscapes of deep neural networks have a great impact on their training, but they have mainly been studied theoretically due to computational constraints. This work vastly reduces the time required to compute such loss landscapes, and uses them to study winning lottery tickets found via iterative magnitude pruning. We also share results that contradict previously claimed correlations between certain loss landscape projection methods and model trainability and generalization error.
\end{abstract}

\section{Introduction}
\label{submission}

Deep neural networks (DNNs) are frequently trained using one of the many variants of stochastic gradient descent (SGD). These methods update a network's parameters using the gradients of the loss w.r.t. said parameters. DNNs have many degrees of freedom (i.e. weights), and their objective functions are thus very-high dimensional. For example, ResNet50 has over 23 million trainable parameters. The "loss landscape" is the same number of dimensions as the weight space plus 1, as each possible configuration of the DNN is evaluated for its loss over some number of test examples (i.e. examples not seen during training).

The first few sections of this paper cover the theoretical background surrounding the loss landscape and introduce the specific visualization method used in this work. The loss landscape (also referred to as "loss surface" and "objective landscape") is constructed by calculating the loss of multiple points in the weight space (i.e. different configurations) of a DNN. Later in the paper we introduce the lottery ticket hypothesis (LTH) and iterative magnitude pruning (IMP) by \cite{lth}, and apply the same loss visualizations to winning lottery tickets (WLTs) created using IMP.

\begin{figure}[ht]
% \vskip 0.2in
\begin{center}
\centerline{\includegraphics[width=\columnwidth]{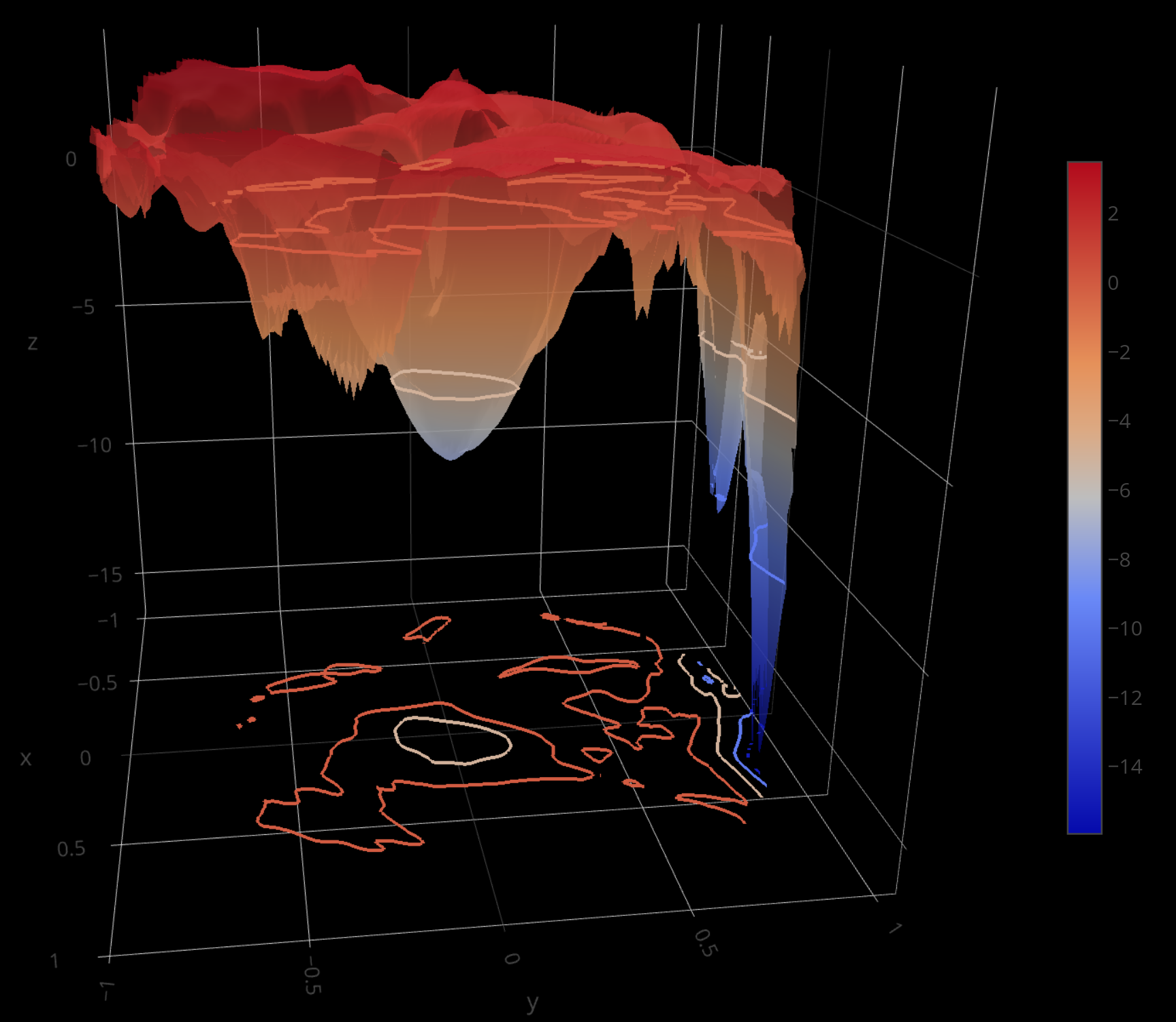}}
\caption{The loss landscape of a ResNet56, with skip connections removed, trained on CIFAR10. Skip connections are a useful inductive bias that makes DNNs both easier to train and achieve lower generalization error. The loss surface is extremely chaotic and non-convex, which hints at some of the difficulty in training these types of deep networks. Note that the z-component is on a natural log scale. View and handle the data yourself \href{http://rkbain.com/loss/\#1}{here}.}
\label{icml-historical}
\end{center}
\vskip -0.2in
\end{figure}

All figures in this paper contain a hyperlink in the caption to view the same data with $Loss Plot$ \cite{bain_loss_1}, an in-browser application built specifically to visualize these types of surface plots. The projected contours, radius based clippings, and other settings can be manually controlled from $Loss Plot$. It is designed to be used with a mouse and keyboard.

\section{Contributions}

\begin{itemize}
\item A method of computing the loss landscape that is 100x faster than previous methods. 
\item Demonstrate the effect of varying batch size on the loss surface of DNNs. Our results contradict previously noted correlations of certain visualization features with trainability and generalization error. 
\item Contrast the loss landscape of randomly pruned DNNs and winning lottery tickets produced by iterative magnitude pruning.
\end{itemize}

\section{Loss Visualization Background}

\begin{figure*}[tb]
  \centering
    \includegraphics[width=15cm]{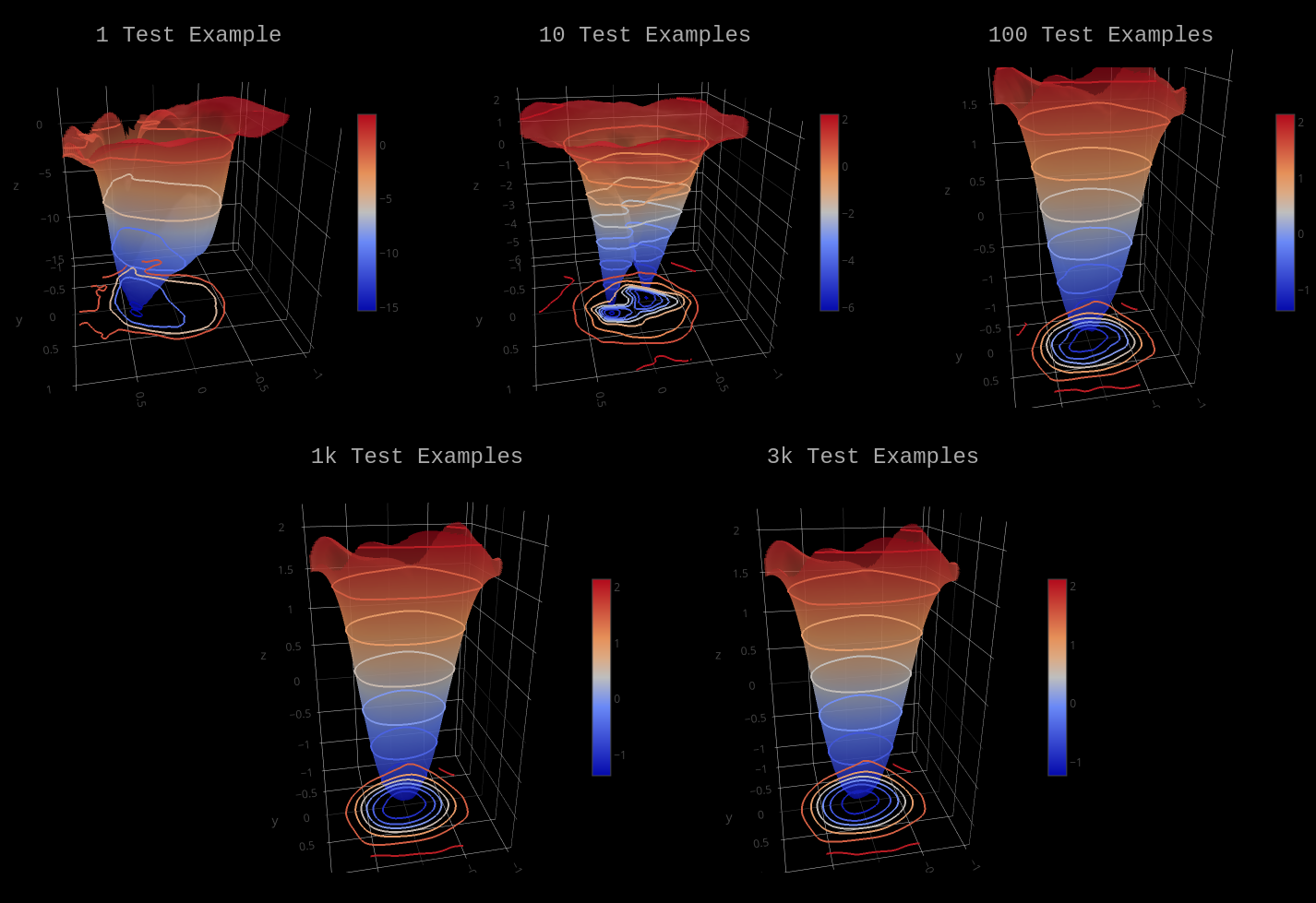}
  \caption{The loss landscape of a ResNet56 trained on CIFAR10. The only difference between each subplot is the number of test examples used to calculate each configuration's generalization loss. The same random directions were used across all subplots to ensure a fair comparison. You can handle the data yourself by clicking \href{https://rkbain.com/loss/\#1}{this link}.}
  \label{resnet}
\end{figure*}

The ability to train DNNs can often be surprising given that they have highly non-convex structures in their objective landscapes. \cite{generalization_measure} explore heuristics for characterizing a DNN's ability to generalize and their best finding was a combination of sharpness and weight magnitude. Flatness is the size of the connected region with similar loss, and sharpness is the antithesis. Flatter minimizers will have larger connected regions of similar loss around the trained minimizer at the center of each surface plot (i.e. there will be little relative curvature around the trained solution). Sharp weight vectors on the other hand have deep crevices immediately surrounding the solution in the loss space.

Modern DNNs often use rectified activation functions (e.g. ReLU) which introduce redundancies in the weight space of models. Scaling one layer by a constant and then scaling the next layer by the inverse results in the same function. Combining batch norm with rectified activation functions has a similar effect, because everything is scaled to a common norm, so a relative comparison between different trained solutions becomes difficult \cite{sharp_flat_symmetries}. Both of these introduce what are referred to as "scaling invariance".

All visualizations of the loss landscape are generated using the methods from \cite{loss_main}. Their open-sourced pytorch code evaluates a 2d grid slice of the weight space centered around the trained minimizer. There are many 2d grid slices to choose from given the dimensionality of modern DNNs. \cite{loss_main} uses a dimensionality reduction technique to choose the slice of the weight space to plot its loss. It begins with creating two random weight vectors by sampling a Gaussian distribution $\mathcal{N}$(0, 1). 

To fix the problems caused by scaling invariance, \cite{loss_main} use "filter-wise normalized directions". There Gaussian sampled direction vectors of length 1 are scaled along the components relating to convolutional neural network (CNN) filters and that subset of the weights norm. I.e. they scale the component subsets of the random directions using the corresponding magnitude of filters:

\begin{center}
$d_{i, j} \leftarrow \frac{d_{i, j}}{\left\|d_{i, j}\right\|}\left\|\theta_{i, j}\right\|$
\end{center}

$d$ represents 1 of the 2 random directions and $\theta$ are the DNN's weights. $d_{i, j}$ represents the j-th filter of the i-th layer of $d$. $\|\cdot\|$ is the Frobenius norm \cite{loss_main}. Filters are not just limited to CNNs. A fully connected layer's filters are the weights connected to one neuron in the resultant layer.

\begin{figure*}[tb]
  \centering
  \includegraphics[width=15cm]{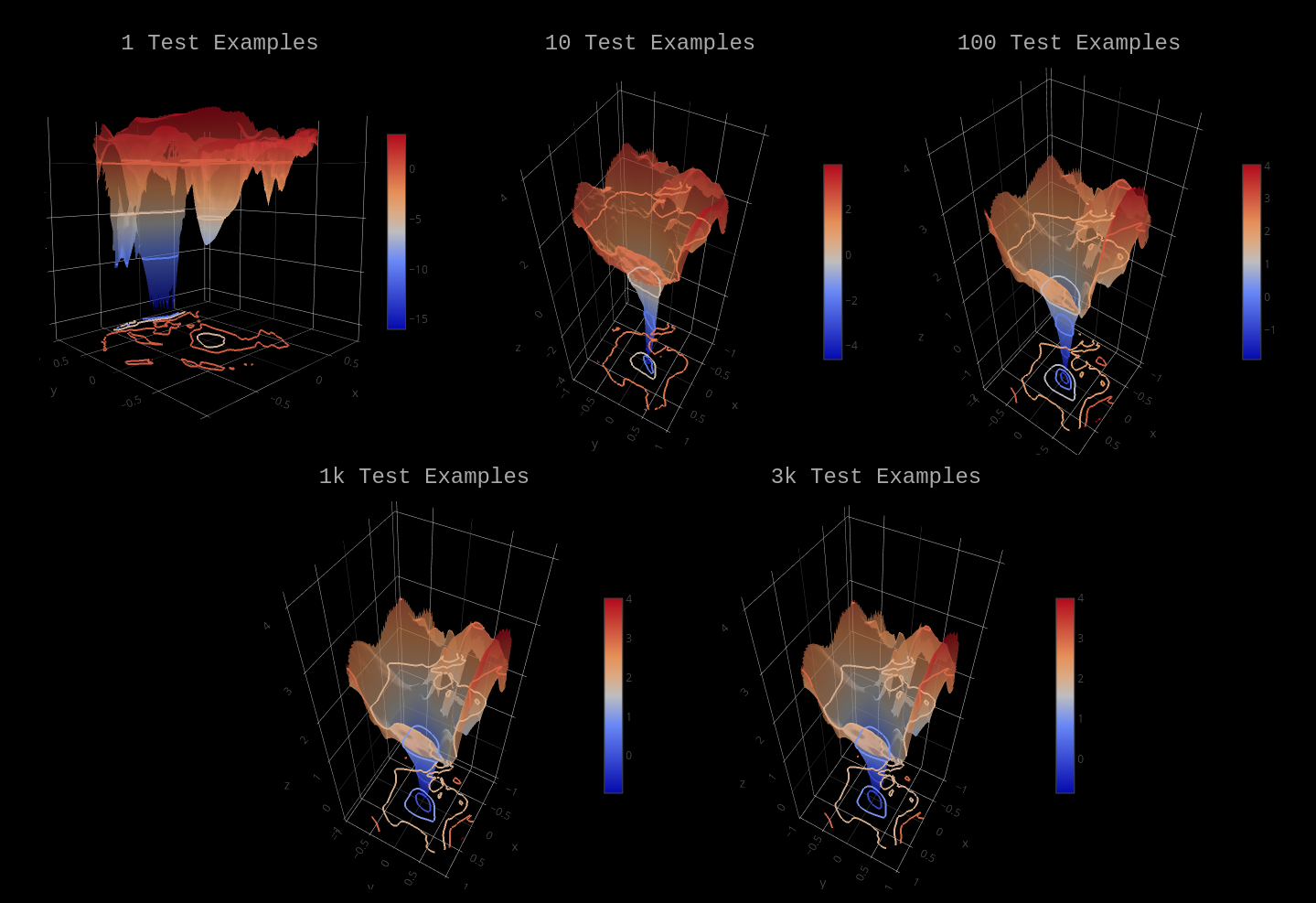}
  \caption{The same model architecture as in \autoref{resnet} but trained without skip connections. The number of examples required to get a general notion of the overall shape is similar, but the landscape has become far more chaotic \& non-convex.}
  \label{noshort}
\end{figure*}

Filter-wise normalization with random directions captures the weighted average of the principle curvatures (i.e. eigenvalues of the Hessian matrix) and enables visual comparison of loss surfaces around different trained solutions, even among different architectures \cite{loss_main}. \cite{loss_main} claim that it also causes the flatness of minimizers to negatively correlate with generalization error. This is contested later in this paper.

\cite{loss_main} reported that it took hours to produce some of their plots using multi-GPU machines. They did not report on the resolution of their plots, but we were able to produce all of this work's 22 individual surface plots with a resolution of 125 x 125 in 5 hours and 30 minutes using a single K40 GPU. 

\section{Creating Loss Visualizations Faster}

The full test set was used previously to produce these types of loss surface plots, yet the general shape of the graph does not change much after a couple hundred validation examples. 

\autoref{resnet} shows the effect of varying the number of test examples to use during evaluation of each specific weight vector. The network being evaluated is a ResNet50 \cite{resnet} trained on CIFAR10. The overall shape changes a lot for the few first examples. By 10 examples the shape has already taken form, and going from 1k to 3k test examples has little effect on the plot. 

This novel observation can realistically lead to a 100x speedup over contemporary methods, even more for large datasets. For example, all surface plots in this paper used a copious random sample of 250 test set images, where CIFAR10 has 10k examples in its test set. This leads to a 40x reduction in computation for this dataset.

\autoref{noshort} shows the same evaluation example sweep but using the released ResNet56 with no skip connections (i.e. noshorts). From the first example the chaotic and non-convex nature of the loss landscape is already apparent. The dominant curvatures of the objective landscape surrounding the trained minimizers is vastly different with and without skip connections.

% *Talk about resnet and how the residual connection is an inductive prior worth having. Heard that python book write about how it might be easier for net's to learn closer to the identity function? The residual *

\begin{figure*}[tb]
  \centering
  \includegraphics[width=15cm]{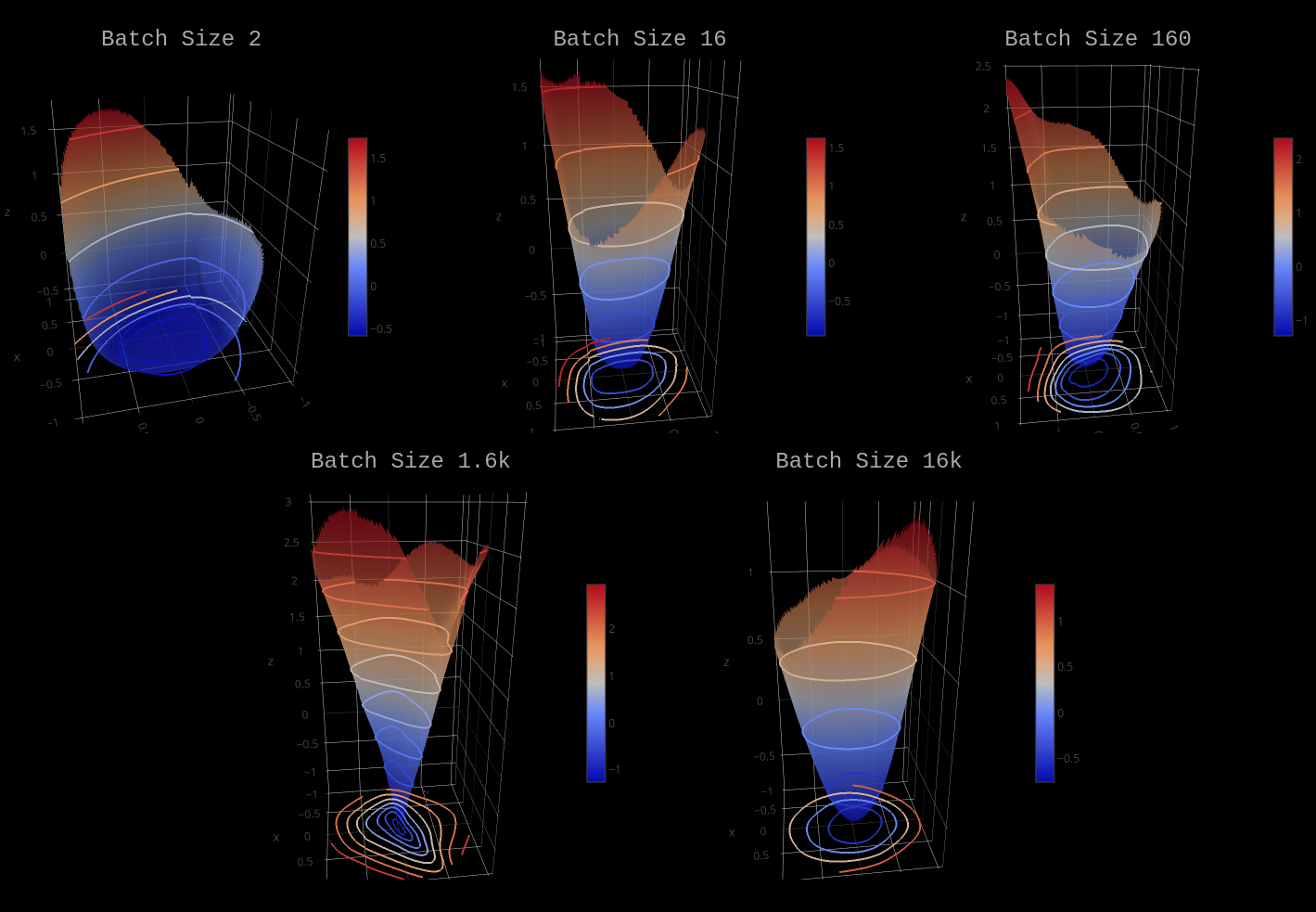}
  \caption{Low batch sizes produce shallow and flat minimizers. The solutions' landscapes become progressively deeper and sharper as the batch size increases. Eventually the trend reverses and the minimizers become shallow and flat again. This experiment's data can be viewed with more detail and control with \href{https://rkbain.com/loss/\#2}{this link}.}
  \label{batchsize}
\end{figure*}

Both ResNet56 models were trained by \cite{loss_main}. They can be obtained from their open-source code \href{https://github.com/tomgoldstein/loss-landscape}{here}. The preceding experiments used different random directions than the original authors in order to verify that the resulting shape is generally consistent across pairs of randomly sampled directions.

\section{Effects of Increasing Batch Size}

\cite{bs_sharp_mins} has hypothesized that larger batch sizes create worse DNNs in part be due to reduced noise in mini-batches, making their weight trajectories more susceptible to falling into and getting stuck in exotic basin structures that result in sharp minimizers. This hypothesis was explored by \cite{loss_main} and the experiment that follows is a natural continuation of theirs.

A LeNet5 \cite{lenet5} architecture with dropout and batch-normalization is trained on FMNIST \cite{fmnist} using Adam \cite{adam} and a variable number of mini-batch sizes. \autoref{batchsize} highlights the effects on the underlying loss surface. Given the faster method of computing these plots we were able to generate 5 high-resolution figures, each with a batch size between 2 and 16,000. The best performance lies between a batch size of 160 and 1,600. It would be interesting to continue increasing the batch size to see if it continually gets worse or if there is a cyclic trend in performance. 

These results contradict the results of \cite{loss_main}, which had previously suggested that these visualizations created correlations between model generalization and the sharpness or flatness of the landscape around the trained minimizer.

\section{Lottery Ticket Hypothesis Background}

The lottery ticket hypothesis (LTH) \cite{lth} states that densely connected DNNs contain sparse sub-network that can exceed the test accuracy of the original network after training on at most the same number of instances. These sparse networks are called winning lottery tickets (WLTs) and do much better than the average random sub-network.

In their work \cite{lth} used Algorithm \autoref{pseudocode_imp} to create winning lottery tickets. The dense network has a mask $m$ applied element-wise to its weights: $m \odot \theta$. The mask starts out as all ones. After training converges the smallest $p$\% of the remaining weights are pruned and their equivalent elements in the mask are set to 0. All remaining weights are re-initialized to their original random values and training begins again. This pattern of training, masking, pruning, and re-setting the weights is repeated until the specified sparsity is reached.

% \begin{figure}[ht]
%     \vskip 0.1in
%     \begin{center}
%     \centerline{\includegraphics[width=\columnwidth]{fu3.png}}
%     \caption{Block diagram showing the difference between the functions learned with and without skip connections. Source: \cite{hands_on_python_ML}}
%     \label{photoshop_combo}
%     \end{center}
%     \vskip -0.1in
% \end{figure}

\begin{algorithm}[tb]
  \caption{Iterative Magnitude Pruning Algorithm}
  \label{pseudocode_imp}
    %\begin{algorithmic}
      \hspace*{0.3cm}1. Train network until convergence.
      \newline
      \hspace*{0.3cm}2. Remove the smallest 10\% of remaining weights per 
      \newline
      \hspace*{0.65cm}
      layer and set them to 0.
      \newline
      \hspace*{0.3cm}
      3. Re-initialize the remaining weights to their original
      \newline
      \hspace*{0.65cm}
      values.
      \newline
      \hspace*{0.3cm}
      4. Repeat steps 2 and 3 until the desired sparsity is met.
    %\end{algorithmic}
\end{algorithm}

% $f\left(x ; \theta_{0}\right)$
% $\theta_{0} \sim \mathcal{D}_{\theta}$
% $f\left(x ; m \odot \theta_{0}^{\prime}\right)$

The following notation is adopted from \cite{lth}:

$P_{m}=\frac{\|m\|_{0}}{|\theta|}$ is the sparsity of $m$, e.g., $P_{m}$ = 25\% when 75\% of weights are pruned.

\section{Loss Landscape of Lottery Tickets}

\begin{figure*}[tb]
  \centering
  \includegraphics[width=15cm]{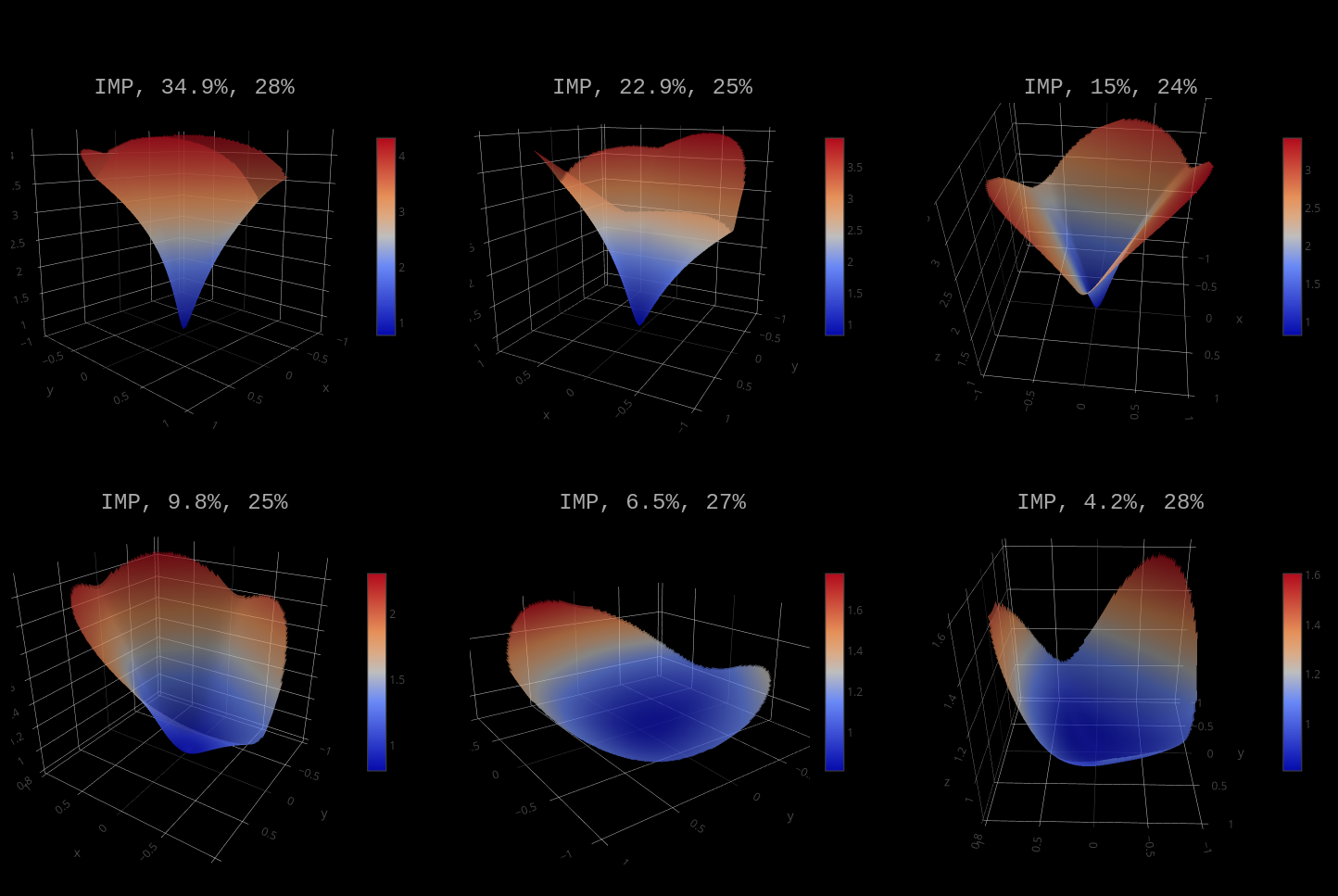}
  \caption{Pictured are the WLTs of LeNet5 trained on CIFAR10. The tickets perform comparably yet the landscapes get flatter and shallower as pruning continues. The subplot titles are tuples of (mask method, $P_{m}$, test accuracy). You can view the data yourself \href{https://rkbain.com/loss/\#3}{here}.}
  \label{imp_masks}
\end{figure*}

\begin{figure*}[tb]
  \centering
  \includegraphics[width=15cm]{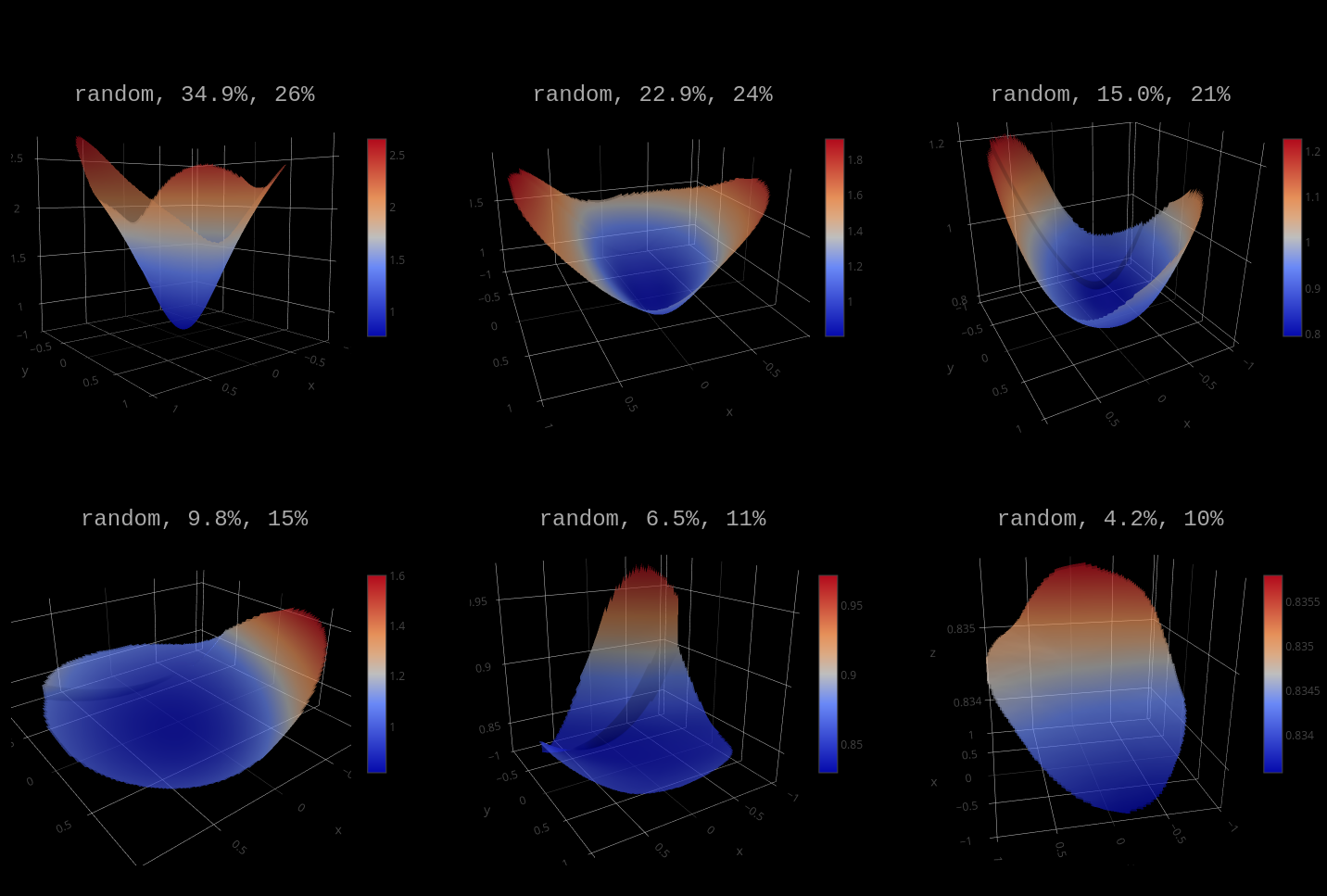}
  \caption{Objective landscapes from random lottery tickets of LeNet5 trained on CIFAR10. These tickets perform much worse relative to the WLTs obtained using IMP. They are also vastly more shallow. The data is accessible from the web using \href{https://rkbain.com/loss/\#3}{this link}.}
  \label{random_masks}
\end{figure*}

\autoref{imp_masks} details how the loss landscape evolves as more and more weights are pruned via Algorithm \autoref{pseudocode_imp}. The LeNet5 architectures are trained on CIFAR10 using an Adam optimizer \cite{adam} and mini-batch size of 9.6k. The best models were obtained using early stopping. After every 35 epochs 10\% of the remaining weights were pruned. A total of 35 pruning iterations occurred, ending with only 3\% of the weights intact. \autoref{random_masks} shows the same progression of $P_{m}$ but using randomly created masks.

\begin{figure}[ht]
    \vskip 0.1in
    \begin{center}
    \centerline{\includegraphics[width=\columnwidth]{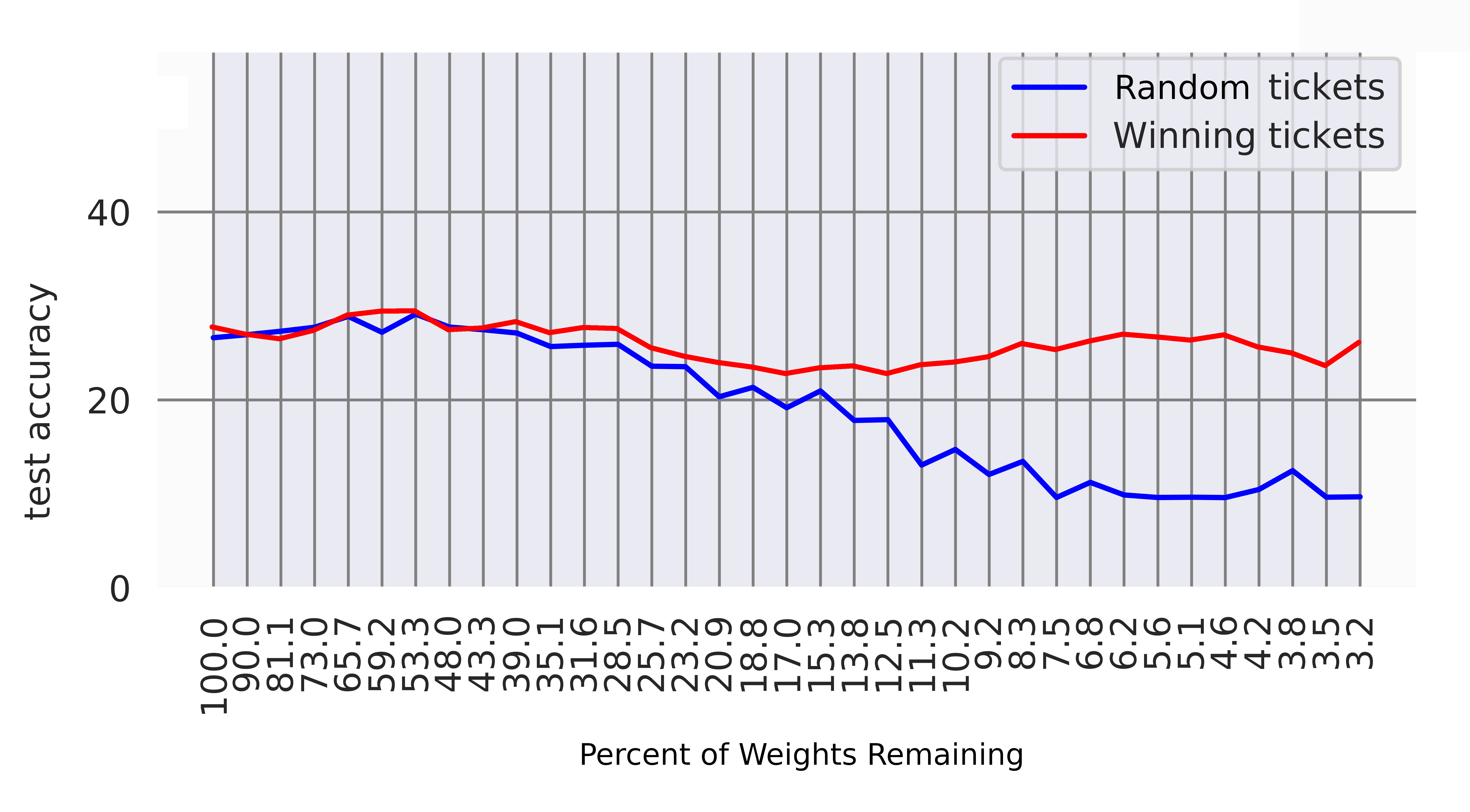}}
    \caption{Training curves of both randomly pruned tickets and tickets derived from Algorithm 1. After around $P_{m}$ = 50\% the gap in performance shows. When $P_{m}$ = 3.2\% the winning ticket is still performing as well as the original dense network and generalizes twice as well as the random ticket of the same sparsity.}
    \label{photoshop_combo}
    \end{center}
    % \vskip 0.1in
\end{figure}

\autoref{photoshop_combo} shows the learning trends for both WLTs and random tickets. The results become interesting around $P_{m}$ = 50\%. It is here that the performance gap between IMP and random pruning begins to occur. It continues to grow all the way through the pruning iterations. By the end of training the gap in test accuracy is 16\%. Note the scale on the bottom-right subplot of \autoref{random_masks}, where $P_{m}$ = 4.2\%. At first glance the slope might seem non-trivial, but the z-component's range is much lesser on this plot. Relative to the others, this loss surface is very flat. This might be an exotic structure that prevents training. These "gradient deserts" seem to be dominated by near 0 curvature, meaning small gradients and weight updates, essentially stalling training. Independently confirming that this gradient desert exists is suggested by the author of this research.

Flatness and shallowness in randomly pruned ticket's loss landscape projections is associated with poor generalization. The IMP method consistently produced more convex and sharper minimizers relative to random masking. 

\section{Discussion}

WLTs have been demonstrated to exist in many architecture \cite{uber, lth, otra_domains_rl_nlp, rejected} and in different task domains like reinforcement learning (RL) and natural language processing \cite{otra_domains_rl_nlp}. Finding WLTs when less than 50\% of the network is pruned seems trivial, but what pragmatic benefit do they offer to practitioners? Finding these winning tickets still requires a lot of compute and fine-tuning. If the sample efficiency gains of WLTs can be had from less compute and data, this could impact fields like Deep RL where sample efficiency is essentially non-existent. 

\cite{uber} found that even better than IMP is keeping weights whose values change by the greatest magnitude, instead of just keeping the largest magnitude weights. \cite{lth} unintentionally foreshadowed this when noting in their Appendix F that winning tickets' weights move further than other weights. \cite{uber} also showed that re-initializing the weights is not as important as retaining the original signs of the weights, lending even more evidence to the idea that re-initializing to the original values is not vital to finding WLTs.

\cite{dont_reinit} revised some of the experiments from \cite{lth} on pruning larger networks while training on ImageNet. This domain usually requires a more exhaustive hyper-parameter search to discover WLTs, but when discovered the results tend to more impressive \cite{lth}. \cite{dont_reinit} demonstrate that using the more de facto training regime with a larger learning rate and momentum SGD instead of Adam produces better results than even the winning tickets, making LTH even less practical. 

\cite{rejected} deduced that 1 hidden-layer neural networks (with some other assumptions about i.i.d. sampling) should have an enlarged convex region appear around a guaranteed optimal solution when pruned correctly. It is difficult to make sense of this work's results in terms of that hypothesis, but it might be an interesting future research direction.

\vskip 25.1in

As originally noted by \cite{lth} their IMP method produces sparse networks in such a way that GPUs do not benefit. The sparsity allows more compression, but not faster inference. In part, this is due to the asynchronous checkpoints that occur while GPUs process matrix calculations in parallel. Multiplications by 0 in a certain batch might be faster (i.e. fewer clock cycles) but the slowest multiplication in that batch of operations has to finish before more calculations can be queued.

\cite{loss_main} experiments using PCA and found that the majority of variance in weight updates lies in only a few dimensions. Perhaps these few very important dimensions are preserved with IMP. It could also be the case that larger learning rates could dislodge the weights from a gradient desert and find more amenable points in the weight space, allowing the recovery of useful training for even randomly masked networks.

The filter-wise normalized random directions method of visualizing the loss landscape has been shown twice not correlating flatness or sharpness with generalization error. However, the convexity of the visualization still seems to correlate with trainability and test performance when the minimizer is deep enough.

\vskip 25.1in 

\textcolor{white}{.}

\vskip 25.1in 

%%% ====== Cite a figure
% \autoref{fig:lossplot}

%%% ============== SUBSECTION HEADER
% \subsection{SUBSECTION}

%%% ============== BOLDFACE TEXT @ BEGINNING OF PARAGRAPH
%\textbf{Paper Deadline:} The deadline for paper submission that is

%%% ================== PSEUDO-CODE BLOCK
% \begin{algorithm}[tb]
%   \caption{Bubble Sort}
%   \label{alg:example}
% \begin{algorithmic}
%   \STATE {\bfseries Input:} data $x_i$, size $m$
%   \REPEAT
%   \STATE Initialize $noChange = true$.
%   \FOR{$i=1$ {\bfseries to} $m-1$}
%   \IF{$x_i > x_{i+1}$}
%   \STATE Swap $x_i$ and $x_{i+1}$
%   \STATE $noChange = false$
%   \ENDIF
%   \ENDFOR
%   \UNTIL{$noChange$ is $true$}
% \end{algorithmic}
% \end{algorithm}

%%% ============== SIMPLE TABLES!
% \begin{table}[t]
% \caption{Classification accuracies for naive Bayes and flexible
% Bayes on various data sets.}
% \label{sample-table}
% \vskip 0.15in
% \begin{center}
% \begin{small}
% \begin{sc}
% \begin{tabular}{lcccr}
% \toprule
% Data set & Naive & Flexible & Better? \\
% \midrule
% Breast    & 95.9$\pm$ 0.2& 96.7$\pm$ 0.2& $\surd$ \\
% Cleveland & 83.3$\pm$ 0.6& 80.0$\pm$ 0.6& $\times$\\
% Glass2    & 61.9$\pm$ 1.4& 83.8$\pm$ 0.7& $\surd$ \\
% Credit    & 74.8$\pm$ 0.5& 78.3$\pm$ 0.6&         \\
% Horse     & 73.3$\pm$ 0.9& 69.7$\pm$ 1.0& $\times$\\
% Meta      & 67.1$\pm$ 0.6& 76.5$\pm$ 0.5& $\surd$ \\
% Pima      & 75.1$\pm$ 0.6& 73.9$\pm$ 0.5&         \\
% Vehicle   & 44.9$\pm$ 0.6& 61.5$\pm$ 0.4& $\surd$ \\
% \bottomrule
% \end{tabular}
% \end{sc}
% \end{small}
% \end{center}
% \vskip -0.1in
% \end{table}

\bibliography{main}
\bibliographystyle{icml2021}

\end{document}